\pdfoutput=1
\documentclass[letterpaper]{article} 
\usepackage{aaai2026}  
\usepackage{times}  
\usepackage{helvet}  
\usepackage{courier}  
\usepackage[hyphens]{url}  
\usepackage{graphicx} 
\usepackage{natbib}  
\usepackage{caption} 
\usepackage{algorithm}
\usepackage{algorithmic}

\usepackage{graphicx}
\usepackage{multirow}
\usepackage{amsmath}
\usepackage{array}
\usepackage{colortbl}
\usepackage{algorithm,algorithmic}
\usepackage{enumitem}
\usepackage{makecell}
\usepackage{float}
\usepackage{mathrsfs}
\usepackage[mathscr]{eucal}
\usepackage{booktabs} 
     
\usepackage{subcaption}
\usepackage{bm}
\usepackage{accsupp}
\usepackage{placeins}
\usepackage{cuted}
\usepackage{amsfonts}
\usepackage{xcolor}
\usepackage{enumitem} 

\usepackage{newfloat}
\usepackage{listings}
\DeclareCaptionStyle{ruled}{labelfont=normalfont,labelsep=colon,strut=off} 
\floatstyle{ruled}
\newfloat{listing}{tb}{lst}{}
\floatname{listing}{Listing}
\title{Towards Closed-Loop Embodied Empathy Evolution: Probing LLM-Centric Lifelong Empathic Motion Generation in Unseen Scenarios}
\author{
    Jiawen Wang\textsuperscript{\rm 1}, Jingjing Wang\textsuperscript{\rm 1}\thanks{Corresponding Author: Jingjing Wang}, Tianyang Chen\textsuperscript{\rm 1}, Min Zhang\textsuperscript{\rm 2}, Guodong Zhou\textsuperscript{\rm 1}
}
\affiliations{
    \textsuperscript{\rm 1}School of Computer Science and Technology, Soochow
University, Suzhou, China\\
\textsuperscript{\rm 2}Harbin Institute of Technology, Shenzhen, China\\
jwwang1031@stu.suda.edu.cn, djingwang@suda.edu.cn, 20244227061@stu.suda.edu.cn, \{mzhang, gdzhou\}@suda.edu.cn

}

\usepackage{bibentry}

\begin{document}

\maketitle


\begin{abstract}
In the literature, existing human-centric emotional motion generation methods primarily focus on boosting performance within a single scale-fixed dataset, largely neglecting the flexible and scale-increasing motion scenarios (e.g., sports, dance), whereas effectively learning these newly emerging scenarios can significantly enhance the model’s real-world generalization ability. Inspired by this, this paper proposes a new \textbf{L}LM-Centric \textbf{L}ifelong \textbf{E}mpathic \textbf{M}otion \textbf{G}eneration (L$^2$-EMG) task, which aims to equip LLMs with the capability to continually acquire emotional motion generation knowledge across different unseen scenarios, potentially contributing to building a closed-loop and self-evolving embodied agent equipped with both empathy and intelligence. Further, this paper poses two key challenges in the L$^2$-EMG task, i.e., the emotion decoupling challenge and the scenario adapting challenge. To this end, this paper proposes an \textbf{E}motion-Transferable and \textbf{S}cenario-Adapted \textbf{M}ixture \textbf{o}f \textbf{E}xperts (ES-MoE) approach which designs a causal-guided emotion decoupling block and a scenario-adapted expert constructing block to address the two challenges, respectively. Especially, this paper constructs multiple L$^2$-EMG datasets to validate the effectiveness of the ES-MoE approach. Extensive evaluations show that ES-MoE outperforms advanced baselines.
\end{abstract}



\section{Introduction}
\begin{figure*}
     \centering
     \includegraphics[width=\textwidth]{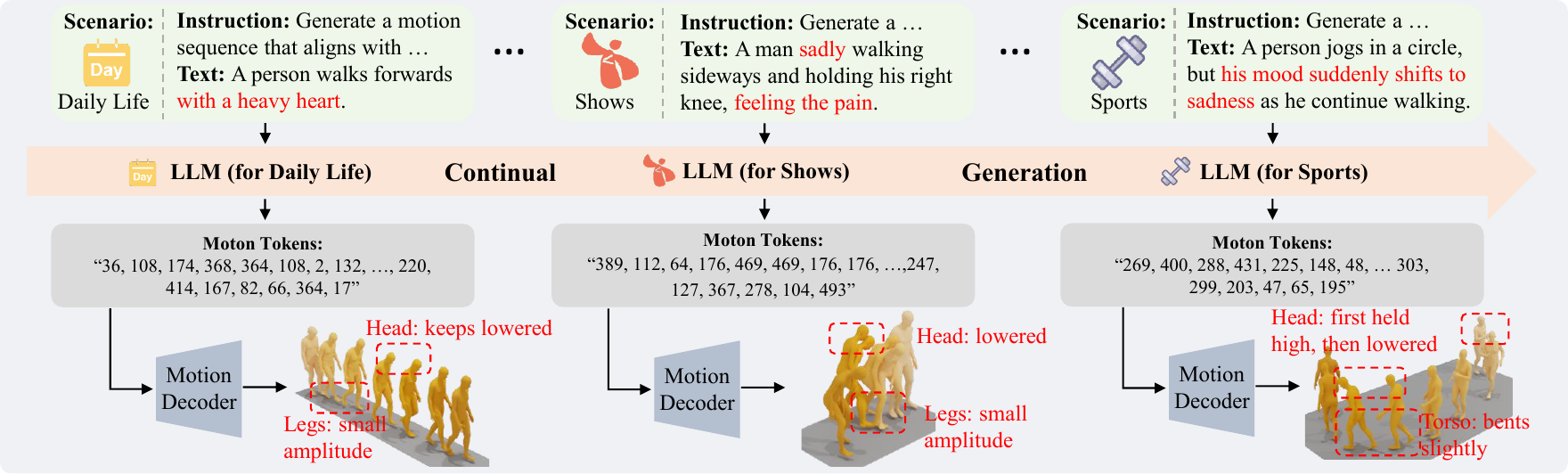}
    \caption{Examples of three scenarios to illustrate our L$^2$-EMG task for continual generation. Red words indicate emotional information in the input, while red boxes highlight the expected emotional motion expressions for the shared emotion \emph{Sad}.}
    \label{fig:example}
\end{figure*}

Human motion generation technology focuses on generating either 3D static motions (i.e., poses \cite{posegpt}) or 3D sequential motions (i.e., motions \cite{DBLP:conf/mm/GuoZWZSDG020,DBLP:conf/iccv/MaoLSL19,DBLP:conf/iclr/TevetRGSCB23}) based on condition signals, which mainly include text \cite{DBLP:conf/cvpr/GuoZZ0JL022,DBLP:conf/iclr/TevetRGSCB23,abs-2301-06052} and speech \cite{DBLP:conf/cvpr/TsengCL23,DBLP:conf/aaai/LiZZS22}, and has broad applications in virtual reality, the metaverse, and game development. Recently, some works \cite{tyumm2024,lifeifei} begin to consider introducing emotional information into the motion generation process due to its potential applications in the field of empathic robots and emotional virtual avatars. However, due to the reliance on specialized equipment for motion data collection and the complexity of emotion annotation processes, existing human motion generation datasets struggle to achieve rapid dynamic updates that are synchronized with real-world scenarios. This lag in data updates causes deployed models to continuously encounter unseen motion scenarios, leading to issues with model generalization decay. Furthermore, even when datasets are updated in a timely manner, considering limited storage and computational budgets, storing all historical data and repeatedly retraining models remains expensive and almost unfeasible.

With these in mind, this paper proposes a \textbf{L}LM-Centric \textbf{L}ifelong \textbf{E}mpathic \textbf{M}otion \textbf{G}eneration (L$^2$-EMG) task, which enables Large Language Models (LLMs) to continually learn emotional motion generation abilities across different unseen scenarios. It can powerfully contribute to developing an empathetic and intelligent embodied agent with closed-loop self-evolving. Specifically, the L$^2$-EMG task sequentially trains LLM on datasets from different unseen motion scenarios, continually learning the emotional generation for new scenarios while preventing the forgetting of motion generation knowledge learned from previous scenarios. As illustrated in Figure 1, for scenarios \emph{``Daily Life"} and \emph{``Sports"}, the model can accurately generate emotional motions that align with the descriptions \emph{``walk"} and \emph{``jog"}, while consistently expressing their shared emotion label \emph{``Sad"}. In this paper, we believe that this new task at least faces two key challenges, which are illustrated as follows.

On the one hand, ensuring the sustainable transfer of emotional representation commonality across scenarios during lifelong learning is challenging, namely the emotion decoupling challenge. The diversity of emotions is not limited to the singularity of a scenario; that is, the motion in each individual scenario contains a variety of emotions. Thus, the model needs to effectively decouple emotional representations with cross-scenario transferability while learning the motion generation style of a specific scenario. As illustrated in Figure 1, two motions from different scenarios—\emph{``Shows"} (training) and \emph{``Sports"}(unseen)—both express the \emph{``Sad"} emotion. We expect the model to capture invariant emotional features (e.g., reduced limb movement and lowered head) from scenario \emph{``Shows"} and transfer them to scenario \emph{``Sports"}. Therefore, this paper believes that a well-behaved approach should be able to decouple emotional representation commonality during cross-scenario lifelong learning and transfer it to new scenarios.

On the other hand, ensuring that the uniqueness of each motion scenario is not forgotten during cross-scenario lifelong learning is also challenging, namely the scenario adapting challenge. Human motion types are diverse and can be categorized into various scenarios, so the model needs to learn the personalized expressions of motion in each scenario while forgetting as little as possible when learning new scenarios. For example, Figure 1 shows motions from three different scenarios. In the second scenario, the \emph{``Shows"} scenario expresses emotions in a more exaggerated manner, while in the third scenario, the \emph{``Sports"} scenario conveys emotions through more professional and skill-oriented movements.  Therefore, this paper believes that a better-behaved approach should be able to capture the uniqueness of each scenario during cross-scenario lifelong learning and avoid catastrophic forgetting when learning new scenarios.

To address the above challenges, this paper proposes an \textbf{E}motion-Transferable and \textbf{S}cenario-Adapted \textbf{M}ixture \textbf{o}f \textbf{E}xperts (ES-MoE) approach, which enables cross-scenario lifelong learning and emotion-enriched human motion generation. Specifically, inspired by recent causal decoupling works \cite{casual1,casual2}, ES-MoE designs a causal-guided emotion decoupling block to decouple and highlight the common emotional representations shared across different motion scenarios. Then, ES-MoE designs a scenario-adapted expert constructing block based on the MoE \cite{DBLP:journals/neco/JacobsJNH91,boost} architecture to learn scenario-specific expressions and facilitate efficient knowledge transfer. Furthermore, we construct multiple L$^2$-EMG datasets to validate the effectiveness of ES-MoE. Comprehensive experiments demonstrate that ES-MoE achieves significant improvements compared to the advanced lifelong learning baselines.

\section{Related Work}

\textbf{Human Motion Generation.} Human motion generation produces diverse and realistic 3D motions from various controls, such as text \cite{abs-2301-06052,DBLP:journals/corr/abs-2312-00063}, audio \cite{DBLP:conf/cvpr/TsengCL23,DBLP:conf/aaai/LiZZS22}, pose \cite{DBLP:journals/pami/Liu0JJLL023,DBLP:conf/iccv/MaoLSL19}, and trajectories. Earlier works \cite{DBLP:conf/iccv/GhoshCOTS21,DBLP:conf/cvpr/GuoZZ0JL022,DBLP:conf/eccv/TevetGHBC22} build shared latent spaces to learn input–motion relations, while recent diffusion-based approaches \cite{DBLP:conf/cvpr/ChenJLHFCY23,DBLP:conf/iccv/ZhangGPCHLYL23} (e.g., MotionDiffuse \cite{DBLP:journals/pami/ZhangCPHGYL24}, MDM \cite{DBLP:conf/iclr/TevetRGSCB23}, EDGE \cite{DBLP:conf/cvpr/TsengCL23}) further improve motion quality. Other works \cite{abs-2301-06052,DBLP:conf/aaai/ZhangHLTLC00YO24,JiangCLYYC23,yang2024unimumo} explore discrete motion representations via VQ-VAE \cite{DBLP:conf/nips/OordVK17}. Recent efforts incorporate emotion understanding \cite{tyumm2024,lifeifei} for emotion-controllable generation. Unlike them, we propose a new L$^2$-EMG task to enhance lifelong learning in T2M models across expanding motion scenarios, enabling intelligent and empathetic embodied agents.

\begin{figure*}
     \centering
     \includegraphics[width=\textwidth]{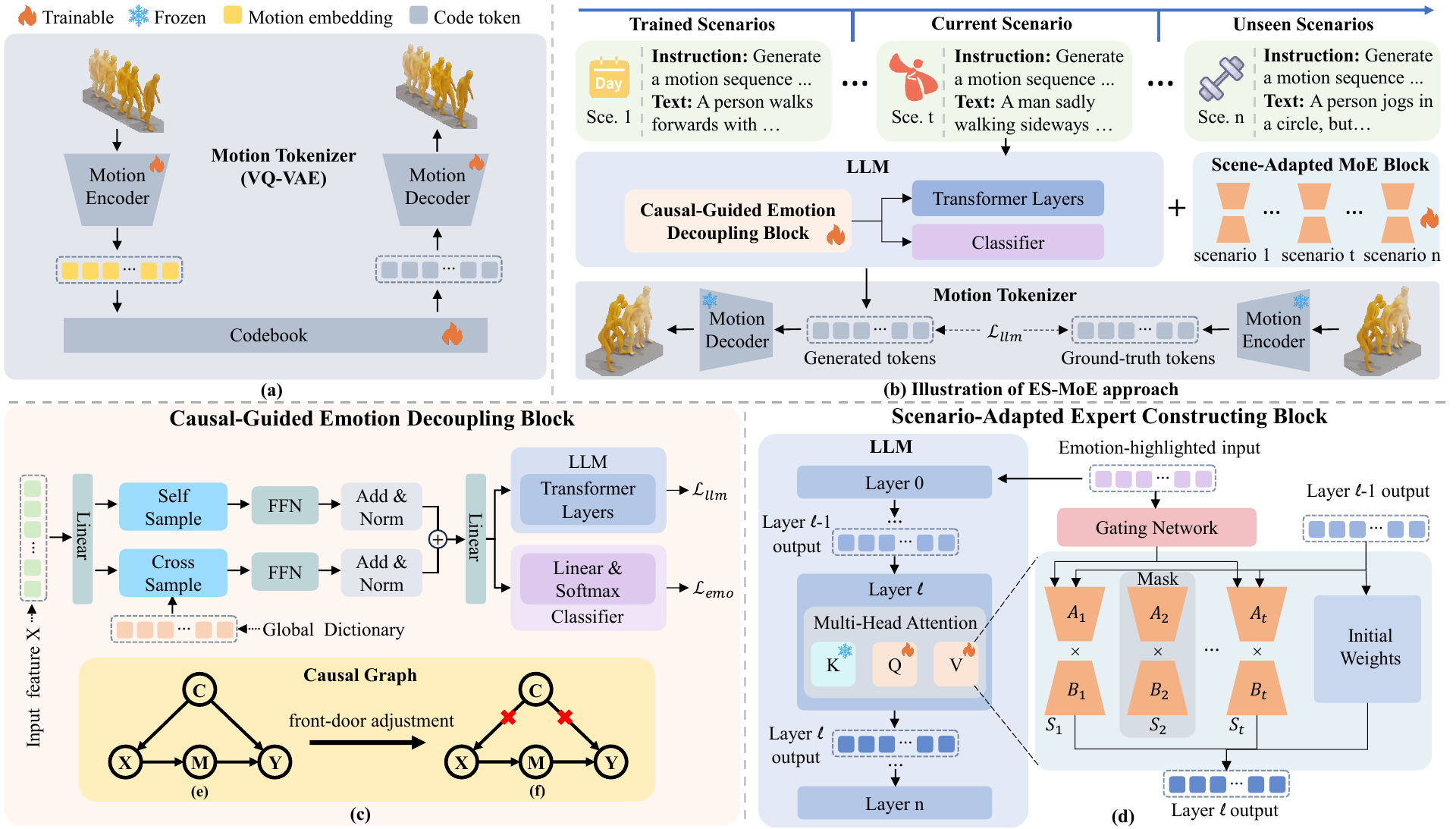}
    \caption{ES-MoE overview: (a) Motion Tokenizer training phase; (b) overall architecture of our approach; (c)/(d) causal-guided emotion decoupling block and scenario-adapted expert constructing block, respectively; (e)/(f) causal intervention graph.}
    \label{fig:model}
\end{figure*}

\textbf{Lifelong Learning.} Lifelong (continual) learning improves future-task generalization while mitigating catastrophic forgetting. Existing methods include parameter regularization \cite{DBLP:conf/nips/Lopez-PazR17,DBLP:conf/aaai/IseleC18, zh}, replay \cite{DBLP:journals/corr/KirkpatrickPRVD16,DBLP:conf/aaai/IseleC18}, and model expansion \cite{mola,sapt}. Regularization preserves past knowledge (e.g., EWC \cite{DBLP:journals/corr/KirkpatrickPRVD16}); replay rehearses stored samples; and expansion methods \cite{o-lora,sapt} allocate task-specific parameters, with SAPT \cite{sapt} enabling selective sharing. MoE-based architectures \cite{DBLP:journals/neco/JacobsJNH91,boost} have recently shown strong potential in continual learning.
Building on these insights, we introduce ES-MoE for the L$^2$-EMG task, which incorporates a causal-guided emotion decoupling block and a scenario-adapted expert construction block to continually learn emotional motion generation across diverse scenarios.

\section{Approach}
In this paper, we propose the L$^2$-EMG task, which aims to enable the model to retain and utilize the emotional motion generation abilities learned from different scenarios and generate natural emotional motions that are appropriate for the new scenario, even when facing entirely new, unseen motion scenarios. Our L$^2$-EMG task is formulated as follows: $\{S_1, S_2, \dots, S_n\}$ represent $n$ different motion scenarios, and $\{D_1, D_2, \dots, D_n\}$ represent the corresponding datasets for these scenarios, where each dataset $D_i=\{(s^i,t_j^i,\mathbf{mt}_j^i)\}_{j=1}^{N_i}$ has the size of $N_i$ and contains scenario-specific instructions $s^i$, text descriptions $t_j^i$ of motions, and output motion tokens $\mathbf{mt}_j^i$. We train the model sequentially on the $\{D_1, D_2, \dots, D_n\}$ datasets. At each time step $i$, the model only has access to the $D_i$ dataset, with the goal that the model learns to generate motions for the scenario at time $i$ without forgetting the scenarios learned before time $i$.

To address the L$^2$-EMG task, we propose the ES-MoE framework (Figure \ref{fig:model}). It contains a Motion Tokenizer that encodes motion sequences into motion tokens, a causal-guided emotion decoupling block that handles emotion decoupling, and a scenario-adapted expert constructing block that enables effective scenario adaptation.

\subsection{Motion Tokenizer}
\label{sec:MT}
To convert 3D human motion data into tokens, we use a VQ-VAE (Vector Quantized Variational AutoEncoder) \cite{DBLP:conf/nips/OordVK17} as the motion tokenizer. As shown in Figure \ref{fig:model} (a), the motion tokenizer consists of two main components: the motion encoder $\mathcal{E}$, which encodes the motion sequence into discrete tokens, and the motion decoder $\mathcal{D}$, which decodes the motion tokens back into the motion sequence.

Specifically, the input to the motion tokenizer is a human motion sequence $\mathbf{mo}$, and the output is the reconstructed motion sequence $\mathbf{mo'}$. At first, motion encoder $\mathcal{E}$ encodes the human motion sequence $\mathbf{mo}$ into the motion feature embedding $\mathbf{e}$. Next, a quantization operation $\mathrm{Quan}(\cdot)$ is applied to transform the motion feature embedding $\mathbf{e}$ into $\mathbf{z}$, a sequence of code vectors, from the learnable codebook $\mathcal{C}=\{\mathbf c_1,\mathbf c_2,\ldots,\mathbf c_k\}$, where $k$ is the size of the codebook and each code vector in the codebook is associated with a discrete token. The quantization operation refers to finding the code vectors in the codebook that are most similar to the motion feature embedding $\mathbf{e}$, which can be mathematically written as:$\mathbf{z}=\{\mathbf z_1,...,\mathbf z_i,...,\mathbf z_{d_\mathbf{e}}\},\mathbf z_i=\mathrm{Quan}(\mathbf e_i)=\mathop{\arg\min} \|\mathbf e_i - \mathbf c_j\|_2$,

where $i\in(1, d_\mathbf{e})$ and $\mathbf c_j \in \mathcal{C}$. $d_\mathbf{e}$ represents the number of columns in the motion embedding $\mathbf{e}$. $\mathbf e_i$ and $\mathbf z_i$ represent the row vectors of $\mathbf{e}$ and its quantized output, respectively. Finally, the decoder $\mathcal{D}$ can decode quantized code tokens into reconstructed motion sequences $\mathbf{mo'}$.

\subsection{Causal-Guided Emotion Decoupling Block}
\label{sec:CE}
In this paper, we leverage the causal intervention technique \cite{book} and design a causal-guided emotion decoupling block to decouple the emotional representation commonality in the motion and further emphasize the commonality of emotion in lifelong learning across different scenarios. This block consists of two parts: the causal intervention graph aiming to decouple emotion via the front-door adjustment strategy \cite{book}, and the deconfounded causal attention that implements this strategy through attention mechanisms in the L$^2$-EMG task, as described below:

\textbf{Causal Intervention Graph for Emotion Decoupling.}
First, to accurately model the causal relation in the decoupling process, we construct a causal intervention graph as shown in Figure \ref{fig:model} (c). Here, X represents the model's input features, including motion and emotion information. M denotes the decoupled feature representation. Y denotes the motion's emotion category, and C represents emotion-irrelevant confounding factors, such as shallow motion semantics. X $\rightarrow$ M $\rightarrow$ Y is a front-door path from X to Y, representing the causal effect of the input features X on the emotion category Y. X $\leftarrow$ C $\rightarrow$ Y is a back-door path from X to Y, representing the causal effect of the confounding factors C on the features X and emotion category Y.

To mitigate the causal influence of emotion-irrelevant confounding factors C and further decouple the emotion information in the features, we adopt a front-door adjustment strategy to alleviate the issue where the model is confused by shallow motion semantics when identifying emotion information. Specifically, we use the do-operator \cite{book} to intervene on X = $x$ to realize the causal effect of X $\rightarrow$ Y and reduce the influence of C on X, thereby blocking the back-door path X $\leftarrow$ C $\rightarrow$ Y. Afterward, we use the front-door adjustment strategy to further compute the causal effect of X $\rightarrow$ M $\rightarrow$ Y ((e) to (f) in Figure \ref{fig:model}), which can be mathematically written as $P({\rm Y}|\mathrm{do}({\rm X}=x))=$:

\begin{equation}
\small
\sum_{m}P({\rm M}=m|{\rm X})\sum_{x}P({\rm X}=x)[P({\rm Y}|{\rm X}=x,{\rm M}=m)]
\label{eq:for}
\end{equation}

\textbf{Deconfounded Causal Attention for Emotion Decoupling.}
Next, we implement the front-door adjustment strategy via the attention mechanism to decouple and highlight emotion information in the input feature X. Specifically, we adopt the Normalized Weighted Geometric Mean (NWGM) \cite{dropout,show} approximation to implement the front-door adjustment strategy in Equation \ref{eq:for}:
\begin{equation}
        P({\rm Y}|\mathrm{do}({\rm X}=x)) \approx {\rm Softmax}(\phi(\mathbf{s}_{x},\mathbf{s}_{m}))
\end{equation}
where $\rm do$ is do-operator \cite{book}. $\phi(\cdot)$ is the parameter network that simulates the predictive distribution $P({\rm Y}|{\rm X}=x,{\rm M}=m)$. $\mathbf{s}_{x}= \sum_{m}P({\rm M}=m|{f(\rm X)})\mathbf{x}$ and $\mathbf{s}_{m} = \sum_{x}P({\rm X}=x|{g(\rm X)})\mathbf{m}$ represent the estimated results of self-sampling and cross-sampling, respectively. $f(\cdot)$ and $g(\cdot)$ are the networks that generate query embeddings for input features X. The generated query embeddings $\mathbf{x}$ and $\mathbf{m}$ correspond to the variables $x$ and $m$, respectively. While self-sampling helps the model decouple emotion information, it remains susceptible to confounding factors. Cross-sampling is thus employed to learn true emotional commonalities from other samples.

We implement the operation of self-sampling and cross-sampling via the attention mechanism. Specifically, the self-sampling result $\mathbf{s}_{x}$ is computed by: $\mathbf{s}_{x} = \mathbf{V_{m}\cdot Softmax(\mathbf{Q_{m}^\top K_{m}})}$, where $\mathbf{Q_{m}}$ comes from $f(\rm X)$. $\mathbf{V_{m}}$ and $\mathbf{K_{m}}$ are obtained by linear transformation of the current input features X. The cross-sampling result $\mathbf{s}_{m}$ is calculated by: $\mathbf{s}_{m} = \mathbf{V_{g}\cdot Softmax(\mathbf{Q_{c}^\top K_{g}})}$, where $\mathbf{Q_{c}}$ comes from $g(\rm X)$. $\mathbf{V_{g}}$ and $\mathbf{K_{g}}$ are obtained from a global dictionary which is compressed from the training set. We perform K-means clustering \cite{kmeans} on the training set samples to initialize this global dictionary. By learning emotion commonality representations from other samples, the cross-sampling result $\mathbf{s}_{m}$ can effectively decouple the emotion information in the input. Finally, we concatenate and integrate $\mathbf{s}_{x}$ and $\mathbf{s}_{m}$ using a Feed-Forward Network (FFN) with parameters $\mathbf{W}$ and $b$, and finally obtain the emotion-highlighted input $\mathbf{h}=\mathbf{W}({\rm FFN}_1(\mathbf{s}_{x})\oplus {\rm FFN}_2(\mathbf{s}_{m}))+b$.

To better decouple the emotional commonality representation, we also design an emotional constraint loss. Specifically, we input the integrated features $\mathbf{h}$ into a simple emotion classification network. This network consists of several fully connected layers, each followed by a ReLU activation function to introduce non-linearity. Finally, the Softmax function outputs the probability distribution of emotional categories. We use the cross-entropy loss function to calculate the difference between the model's predicted emotional category probabilities and the true motion emotion label: $
    \mathcal{L}_{emo} = {\rm CrossEntropy}(y_e, \hat{y}_{e})
$, where $y_e$ and $\hat{y}_{e}$ represent the true motion emotion label and the prediction result of the classification network, respectively.

\begin{figure}[t]
\begin{algorithm}[H]
    \renewcommand{\algorithmicrequire}{\textbf{Input:}}
	\renewcommand{\algorithmicensure}{\textbf{Output:}}
	\caption{ES-MoE for lifelong learning on scenario $i$}\small
    \label{ag}
    \begin{algorithmic}[1] 
        \STATE   \textbf{Input: }Pre-trained mdoel parameters ${\theta}$; Trained expert parameters $\{\Delta \theta_j \mid j = 1, \dots, i-1\}$, i-th expert parameters $\Delta \theta_i$ to be trained
	    \STATE  \textbf{Output: }Final updated model parameters ${\theta'}$
        \STATE Obtain the emotion-highlighted input $\mathbf{h}$ 
        \STATE Construct the $i$-th expert parameters $\Delta \theta_i=\mathbf A_i\mathbf B_i$ for the current scenario based on the LoRA method
        \STATE Calculate the expert weights $W_{i}$ \hfill $\triangleright$ Eq.\eqref{eq:gate}
        \STATE 
        \begin{minipage}[t]{\linewidth}
        Aggregate the parameters of all experts to obtain\\
        \makebox[\linewidth][l]{the final updated model weights $\theta'$ }
        \end{minipage}
    \end{algorithmic}
\end{algorithm}
\end{figure}

\begin{table*}[t]
\scriptsize
\begin{tabular*}{\textwidth}{@{\hskip1em}@{\extracolsep{\fill}}lccccccccc}
\toprule
    \multirow{2}{*}{\textbf{Unseen}} & \textbf{\#Daily Lif}e & \textbf{\#Sports} & \textbf{\#Dance} & \textbf{\#Shows} & \textbf{\#Game} & \textbf{\#Animation} & \textbf{\#Instrument Play} & \textbf{\#Acrobatics} & \textbf{Total} \\
      \cline{2-10}& 5407        & 2191   & 2068       & 1101       & 1618     & 1719    & 3903     & 1909        & 19916  \\
    \midrule[0.3pt] 
    \multirow{2}{*}{\textbf{Mixed}} & \textbf{\#Scenario1} & \textbf{\#Scenario2} & \textbf{\#Scenario3} & \textbf{\#Scenario4} & \textbf{\#Scenario5} & \textbf{\#Scenario6} & \textbf{\#Scenario7} & \textbf{\#Scenario8} & \textbf{Total} \\
      \cline{2-10}& 2489        & 2489   & 2489       & 2489       & 2489     & 2489    & 2489     & 2493        & 19916  \\
\bottomrule
\end{tabular*}
\caption{Statistics of our constructed L$^2$-EMG dataset. Unseen and Mixed denote two ways of splitting our dataset.}
\label{tab:dataset}
\end{table*}

\subsection{Scenario-Adapted Expert Constructing Block}
\label{sec:SME}
Further, we design a scenario-adapted expert constructing block based on the MoE \cite{DBLP:journals/neco/JacobsJNH91,boost} architecture to adapt to the gradually increasing motion scenarios and enable efficient knowledge transfer across different motion scenarios. Algorithm \ref{ag} introduces the process of constructing experts when encountering a new $i$-th motion scenario, and two main processes are detailed as follows:


\textbf{LoRA Experts Construction.}
To enable the model to effectively learn and retain knowledge across different motion scenarios, we design multiple experts for different scenarios based on LoRA \cite{DBLP:conf/iclr/HuSWALWWC22}. Specifically, we train the model sequentially on $n$ different motion scenarios $\{S_1, S_2, \dots, S_n\}$. At each time step $i$, our training objective is:
\begin{equation}
\small
\mathop{\arg\min}\limits_{\theta'}\sum_{(t,\mathbf{mt})\in D_i}\mathcal{L}_{i}(f_{\theta'}(t),\mathbf{mt}), \theta'={\theta,\Delta\theta_{1},...,\Delta\theta_{i}}
\end{equation}
where $D_i$ represents the dataset of $i$-th scenario. $f_{\theta'}$ represents the model with weight ${\theta'}$. $\Delta\theta_{i}$ represents the change in model parameters after training on the $i$-th scenario. 

Given the simplicity and effectiveness of the LoRA method, we use the LoRA module with parameters represented as $\mathbf A_{i} \times \mathbf B_{i}$ to simulate the parameter updates after training on the $i$-th scenario. For each motion scenario $S_{i}$, we assign a LoRA module as the motion generation expert for that scenario. Since motion scenarios are inherently interrelated, we adopt a Mixture of Experts (MoE) architecture to complete the motion generation task corresponding to the $i$-th scenario, which integrates the experts that have already been trained up to time step $i$, rather than solely relying on the experts trained at time step $i$. The model parameters after training at the $i$-th time step can be represented as:
$
    \theta'=\theta + \sum_{j=1}^{i}W_{j}\Delta\theta_{j}=\theta + \sum_{j=1}^{i}W_{j}\mathbf A_j\mathbf B_j
    \label{eq:agg}
$, where $\theta$ represents pre-trained model parameters. $\Delta\theta_{i}=\mathbf A_i\mathbf B_i$ represents the parameters of the LoRA expert corresponding to the $i$-th scenario. $W_{i}$ represents the weight of the $i$-th expert, measuring the expert's contribution.


\textbf{Gating Network Construction.}
Next, we design a gating network ${\rm Gate(\cdot)}$ to generate weight $W$ for each LoRA expert. Specifically, the implementation is divided into three steps: 1) First, we use orthogonal initialization \cite{DBLP:journals/corr/SaxeMG13,DBLP:conf/iclr/PengT0YJ24} to generate a set of mutually orthogonal expert keys, ensuring discrimination between different experts in the key feature space. 2) Then, we process $\mathbf{h}$, the emotion-highlighted input sequence in above Section, through a learnable down and up projection layer \cite{sapt} and map it to a query embedding aligned with the expert key dimension. 3) Finally, we compute the dot product similarity between the query vector and each expert key to generate the weight of each expert, which can be denoted as: 

\begin{equation}
    W_{i}={\rm Gate}(\mathbf{h},\mathbf{K}_{i})=\frac{e^{{\rm Query Embedding}(\mathbf{h})^\top \cdot \mathbf{K}_{i}}}{\sum_{j=1}^{i} e^{{\rm Query Embedding}(\mathbf{h})^\top \cdot \mathbf{K}_{j}}}
    \label{eq:gate}
\end{equation}
where $\mathbf{h}$ represents the emotion-highlighted input. $\mathbf{K}_{i}$ represents the key vector of $i$-th expert.

At time step $i$, only the LoRA matrix and key vector of the $i$-th expert are trained, while the query embedding network remains trainable. To avoid over-reliance on previously trained experts, we randomly mask trained experts from participating in weight assignments depending on the training time point. Among the remaining experts, we select the top k most relevant ones based on their weight sizes and recalculate the weights using the Softmax function.

\subsection{Optimization for ES-MoE}
\label{section:3.4}
Our model training consists of two stages. First, we train a VQ-VAE–based motion tokenizer that encodes motion sequences into tokens and reconstructs the original motion. Second, we sequentially fine-tune the LLM using motion generation datasets from different scenarios.

\textbf{Stage1. Motion Tokenizer Training.} To train VQ-VAE as a motion tokenizer, we follow T2M-GPT to set the optimization goal $\mathcal{L}_{vq}$. $\mathcal{L}_{vq}$ consists of three main components: the motion reconstruction loss $\mathcal{L}_{re}$, the embedding loss $\mathcal{L}_{embed}$, and the commitment loss $\mathcal{L}_{commit}$, which can be denoted as: $\mathcal{L}_{vq}=\mathcal{L}_{re}+\mathcal{L}_{embed}+\mathcal{L}_{commit}$.

\textbf{Stage2. Continual Scenario Tuning.} In the second stage, we construct multiple L$^2$-EMG datasets (detailed in Section 4.1) to endow the model with the ability to continually generate emotional motions across diverse motion scenarios. Then, we sequentially fine-tune the model on these datasets from different scenarios using the instruction ``Generate a motion sequence that aligns with the following emotional text description." The loss in the second stage can be expressed as: $\mathcal{L}=\mathcal{L}_{llm}+\lambda\mathcal{L}_{emo}$. Here, $\mathcal{L}_{llm}$ denotes the LLM next-token prediction loss, and $\mathcal{L}_{emo}$ denotes the emotion classification loss mentioned in Causal-Guided Emotion Decoupling Block, and $\lambda$ controls the weight of $\mathcal{L}_{emo}$.

\section{Experimental Settings}

\begin{table*}[t]
\centering
\label{tab:result}
\footnotesize
\renewcommand{\arraystretch}{1}
\setlength{\tabcolsep}{5.5pt}
\begin{tabular}{l l c c c c c c c c c c c c}
\toprule
\multirow{2}{*}{\textbf{Backbone}} & \multirow{2}{*}{\textbf{Approach}} & \multicolumn{6}{c}{\textbf{Unseen L$^2$-EMG Dataset}} & \multicolumn{6}{c}{\textbf{Mixed L$^2$-EMG Dataset}} \\
\cmidrule(lr){3-8} \cmidrule(lr){9-14}
 & & \textbf{AF}$\downarrow$ & \textbf{AR}$\uparrow$ & \textbf{AD}$\uparrow$ & \textbf{AMM}$\uparrow$ & \textbf{AWF}$\uparrow$ & \textbf{FR}$\downarrow$ & \textbf{AF}$\downarrow$ & \textbf{AR}$\uparrow$ & \textbf{AD}$\uparrow$ & \textbf{AMM}$\uparrow$ & \textbf{AWF}$\uparrow$ & \textbf{FR}$\downarrow$ \\
\hline
\multirow{2}{*}{\makecell{LLaMA2 \\ (Non-CL)}}
& MTL & 1.05 & 0.281 & 9.63 & 1.56 & 0.383 & - & 1.05 & 0.281 & 9.63 & 1.56 & 0.383 & - \\
& SeqLoRA & 3.58 & 0.157 & 8.83 & 1.10 & 0.198 & 6.70 & 2.61 & 0.195 & 9.32 & 1.07 & 0.239 & 5.17 \\
\hline
\multirow{5}{*}{\makecell{LLaMA2 \\ (CL)}} & LwF-LoRA & 4.38 & 0.164 & 8.44 & 0.95 & 0.152 & 5.76 & 3.36 & 0.215 & 8.81 & 1.27 & 0.188 & 3.70 \\
& EPI & 2.21 & 0.180 & \textbf{9.86} & 1.47 & 0.256 & 4.89 & 1.99 & 0.207 & 9.13 & 1.33 & 0.283 & 1.51 \\
& O-LoRA & 2.35 & 0.214 & 9.29 & 1.18 & 0.282 & 2.75 & 2.20 & 0.236 & 9.82 & 1.29 & 0.296 & -0.42 \\
& Prog-Prompt & 7.30 & 0.102 & 6.75 & \textbf{1.84} & 0.128 & 7.16 & 5.77 & 0.153 & 8.12 & \textbf{1.87} & 0.134 & 5.09 \\
& SAPT & 2.12 & 0.237 & 9.61 & 1.59 & 0.313 & -0.54 & 1.65 & 0.245 & 9.47 & 1.82 & 0.327 & -1.97 \\
\hline

\multirow{4}{*}{\makecell{LLaMA2 \\ (Ours)}} & \textbf{ES-MoE} & \textbf{1.89} & 0.241 & 9.74 & 1.47 & \textbf{0.340} & \textbf{-1.03} & \textbf{1.39} & \textbf{0.259} & \textbf{9.87} & 1.65 & \textbf{0.347} & \textbf{-3.03} \\
& w/o CGED & 2.07 & \textbf{0.248} & 9.53 & 1.39 & 0.312 & -0.75 & 1.70 & 0.247 & 9.44 & 1.77 & 0.305 & -1.74 \\
& w/o SAMoE & 2.48 & 0.206 & 9.08 & 1.78 & 0.286 & 3.26 & 1.96 & 0.229 & 9.28 & 1.58 & 0.273 & 0.83 \\
& w/o $\mathcal{L}_{emo}$ & 2.01 & 0.232 & 9.45 & 1.42 & 0.326 & -0.86 & 1.52 & 0.237 & 9.75 & 1.52 & 0.336 & -2.59 \\
\bottomrule
\end{tabular}
\caption{Comparison of our ES-MoE approach with other approaches on the L$^2$-EMG dataset. `$\uparrow$'(`$\downarrow$') indicates that the values are better if the metric is larger (smaller).}

\end{table*}

 \textbf{Dataset Construction.} T\textbf{Dataset Construction.} We build two datasets corresponding to ES-MoE’s two training stages.
\textbf{(1)} For the \textbf{motion tokenizer training} stage, we construct a text–motion pair dataset using EmotionalT2M \cite{tyumm2024} and selected subsets of Motion-X \cite{DBLP:conf/nips/LinZLCZWZ23}. Samples contain an emotional motion description and its motion sequence. Since EmotionalT2M is small, we expand it with Motion-X, whose texts and emotion labels are merged following \cite{tyumm2024} using ChatGLM \cite{DBLP:conf/iclr/ZengLDWL0YXZXTM23}. The processed Motion-X data is then combined with EmotionalT2M.
\textbf{(2)} For the \textbf{continual scenario training} stage, we encode motions into discrete motion tokens via the trained tokenizer and convert them into instruction data. We categorize all samples into eight scenarios based on annotated scenario labels, with two annotators and expert adjudication (Kappa = 0.85).
\textbf{(3)} To evaluate L$^2$-EMG, we design two lifelong learning settings: the \textbf{Unseen L$^2$-EMG} dataset, which sequentially fine-tunes across eight scenario-specific subsets, and the \textbf{Mixed L$^2$-EMG} dataset, which randomly mixes scenarios to mimic real-world incremental data. Each subset has a primary scenario with fewer motions from others and is split into train/val/test (0.8/0.05/0.15). More details are provided in Table \ref{tab:dataset}.

\textbf{Evaluation Metrics.} Following prior studies \cite{DBLP:conf/cvpr/GuoZZ0JL022, tyumm2024}, let $f_{i,j}, r_{i,j}, d_{i,j}, m_{i,j}, wf_{i,j}$ denote the widely-used motion generation metrics, i.e., FID, top-1 R-Precision, diversity, multimodality score, and weighted F1-score of the generated motions in scenario $j$ after the model has been trained on scenario $i$, respectively. Based on these, the evaluation metrics for our ES-MoE are calculated as follows:
\textbf{Average FID (AF)} of all generated motions with 
 different scenarios after training on the final motion scenario $N$. It is computed by $AF_{N}=\frac{1}{N}\sum_{j=1}^Nf_{N,j}$;
\textbf{Average R-Precision (AR):} The average text-motion match top-1 precision of all generated motions with different scenarios after training on the final motion scenario $N$. It is computed by $AR_{N}=\frac{1}{N}\sum_{j=1}^Nr_{N,j}$;
\textbf{Average Diversity (AD)} performance of all scenarios after training on the final motion scenario $N$. It is computed by $AD_{N}=\frac{1}{N}\sum_{j=1}^Nd_{N,j}$;
\textbf{Average MultiModality (AMM)} performance of all scenarios after training on the final motion scenario $N$. It is computed by $AMM_{N}=\frac{1}{N}\sum_{j=1}^Nm_{N,j}$;
\textbf{Average Weight F1-score (AWF)} to evaluate the emotion performance of generated motions proposed by \cite{tyumm2024}. We compute AWF after training on the final motion scenario $N$. It is computed by ${AWF}_{N}=\frac{1}{N}\sum_{j=1}^Nwf_{N,j}$;
\textbf{Forgetting Rate (FR)} of the model on the first $N-1$ motion scenarios after training on the final motion scenario $N$, measuring how much knowledge has been forgotten during the lifelong learning process. It is computed by $F_{N}=\frac{1}{N-1} \sum_{j=1}^{N-1} \left( \max_{k=j}^{N-1} r_{k,j} - r_{N,j} \right)$;

\textbf{Implementation Details and Baselines.} The baselines we choose include: Multi-Task Learning (MTL, which trains a model on multiple tasks simultaneously), SEQ-LoRA (sequentially fine-tuning the model using the LoRA method in a predefined order). LwF-LoRA \cite{lwf}, EPI \cite{epi}, O-LoRA \cite{o-lora}, Prog-Prompt \cite{pp}, and SAPT \cite{sapt}. In the continual scenario tuning stage, we use LLaMA 2 (7B) (llama.com/llama2) as the backbone of all baselines and ES-MoE, and fine-tune it with LoRA for a fair comparison. Following prior works \cite{o-lora}, we compute the results of all CL baselines three times with different scenario orders and take the average as the final score.

\begin{figure*}[t]
     \centering
     \includegraphics[width=\textwidth]{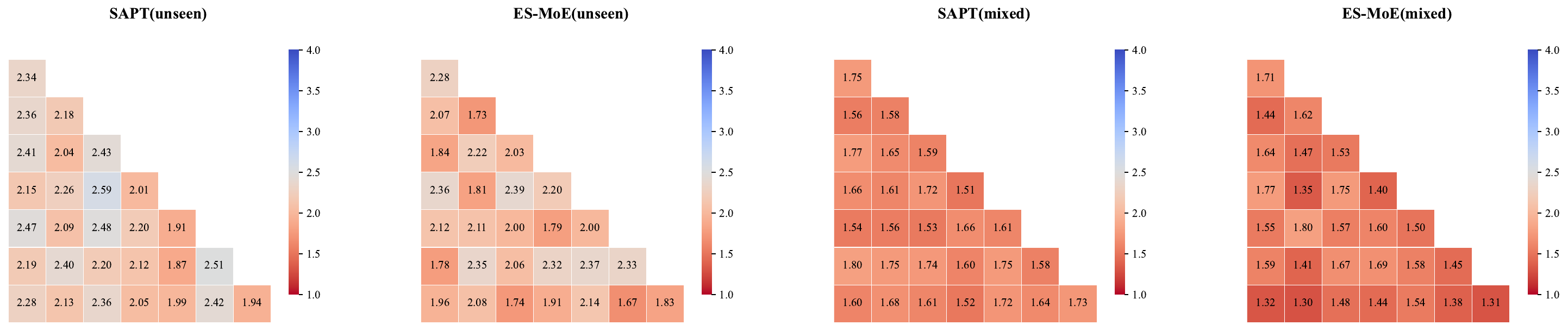}
    \caption{Detailed AF results of four approaches on L$^2$-EMG (Unseen), L$^2$-EMG (Mixed), where bluer represents severe forgetting of generation ability across different motion scenarios, indicating low performance; redder represents strong retention of generation ability across different motion scenarios, indicating high performance.}
    \label{fig:vis}
\end{figure*}

\begin{figure*}
    \centering
    \includegraphics[width=\linewidth]{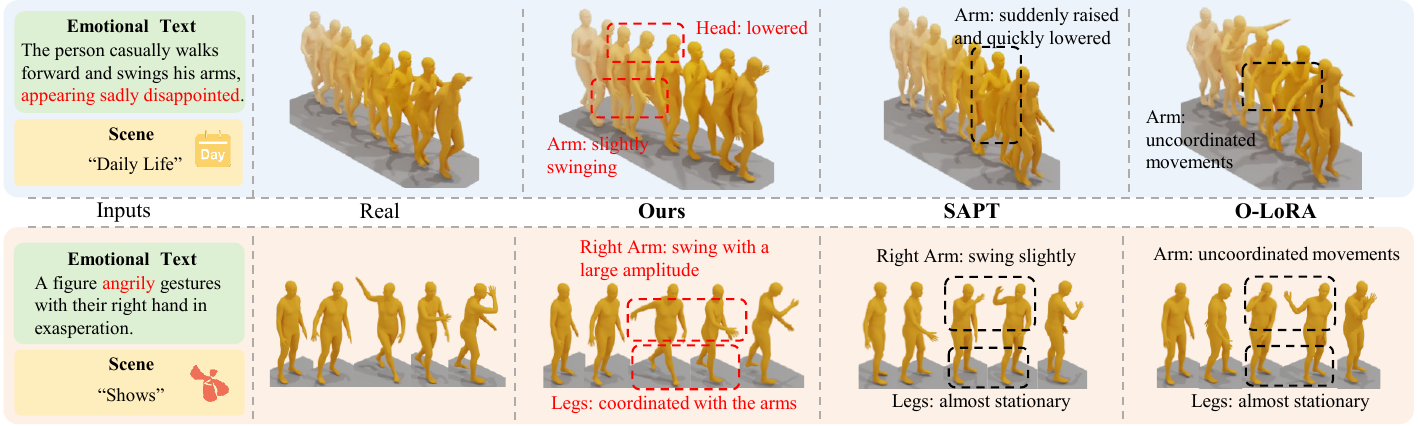}
    \caption{Visualization of motions generated by ES-MoE and other baselines. Red boxes indicate more precise emotional expressions achieved by ES-MoE, whereas black boxes indicate the limitations observed in motions generated by the baselines.}
    \label{fig:qualititave comparision}
\end{figure*}

\section{Results and Discussion}

\textbf{Main Experimental Results.} Table \ref{tab:result} presents a comparative analysis of different approaches on the L$^2$-EMG dataset. From this table, we can see that: \textbf{1)} \textbf{Performance on Unseen L$^2$-EMG Dataset.} On the Unseen L$^2$-EMG Dataset, our ES-MoE approach outperforms other baselines on almost all metrics. For instance, compared to the best-performing baseline, SAPT, ES-MoE achieves better results on the AF, AR, and AD metrics, suggesting that it generates more natural, coherent, and text-aligned motions in diverse scenarios. On the AMM metric, ES-MoE achieves comparable results, indicating that the motions generated by ES-MoE are diverse and rich. Moreover, ES-MoE also achieves better performance in the AWF and FR metrics, further confirming its effectiveness in decoupling emotional information from motions and its flexibility in adapting to entirely new motion scenarios.
\textbf{2)} \textbf{Performance on Mixed L$^2$-EMG Dataset.} On the Mixed L$^2$-EMG Dataset which better aligns with real-world applications, our ES-MoE approach also surpasses other baselines on most metrics. This indicates that ES-MoE is not only effective at decoupling emotional information and generating realistic and natural emotional motions but also adapts well to the real-world scenario.

\textbf{Effectiveness Study for Emotion Decoupling.} To validate ES-MoE’s effectiveness in addressing the emotion decoupling challenge, we conduct ablation studies (Table \ref{tab:result}). Specifically, \textbf{w/o CGED} and \textbf{w/o $\mathcal{L}_{emo}$} denote settings where the Causal-Guided Emotion Decoupling Block (CGED) and the $\mathcal{L}_{emo}$ do not work, respectively. From this table, we can see that: 1) \textbf{w/o CGED} performs worse on both unseen and mixed datasets, especially in AWF score. This justifies the effectiveness of the CGED block in decoupling emotional commonality and enabling the generation of more emotionally expressive motions. 2) \textbf{w/o $\mathcal{L}_{emo}$} also shows a performance drop in AWF and AR on both datasets. This further demonstrates the effectiveness of the CGED block in addressing the challenge of emotional decoupling.

\textbf{Effectiveness Study for Scenario Adapting.} To validate the capability of ES-MoE in addressing the scenario adaptation challenge, we conduct related experiments, as shown in Table 2. Specifically, \textbf{w/o SAMoE} refers to the setting where the Scenario-Adaptive Mixture of Experts(SAMoE) does not work. From this table, we can see that: \textbf{w/o SAMoE} shows a significant performance drop on both types of datasets compared to ES-MoE, with all metrics substantially decreased, especially AF and AR. This demonstrates that scenario-adapted experts can efficiently adapt to newly introduced motion scenarios, justifying the effectiveness of SAMoE in addressing the scenario adapting challenge.

\textbf{Forgetting Analysis across Different Scenarios.} We visualize the FID performance of SAPT (best baseline) and ES-MoE on Unseen and Mixed L$^2$-EMG datasets in Figure \ref{fig:vis}. Each matrix entry (i, j) denotes the FID score on the j-th scenario after training on the i-th scenario. Results show: 1) Final-scenario tests show no clear performance drop—sometimes improvement—indicating minimal forgetting in cross-scenario lifelong learning. 2) SAPT and ES-MoE achieve effective knowledge transfer, with ES-MoE consistently yielding lower FID, reflecting more accurate emotion decoupling and stronger scenario adaptability.

\textbf{Qualitative Analysis via Visualization.} Figure \ref{fig:qualititave comparision} shows visual comparisons of motions generated by ES-MoE and other methods. We observe: 1) All methods capture basic patterns such as walking and arm swinging, but O-LoRA often produces incoherent or uncoordinated motions—for example, in the first case, it swings both arms while walking, causing unnatural movement. 2) SAPT shows limited emotional understanding. In the second case, although it generates a coherent waving motion, the amplitude is too small to convey the intended angry emotion. In contrast, ES-MoE produces coordinated motions that accurately express the target emotion and match the scenario.

\begin{figure*}[t]
     \centering
     \includegraphics[width=\textwidth]{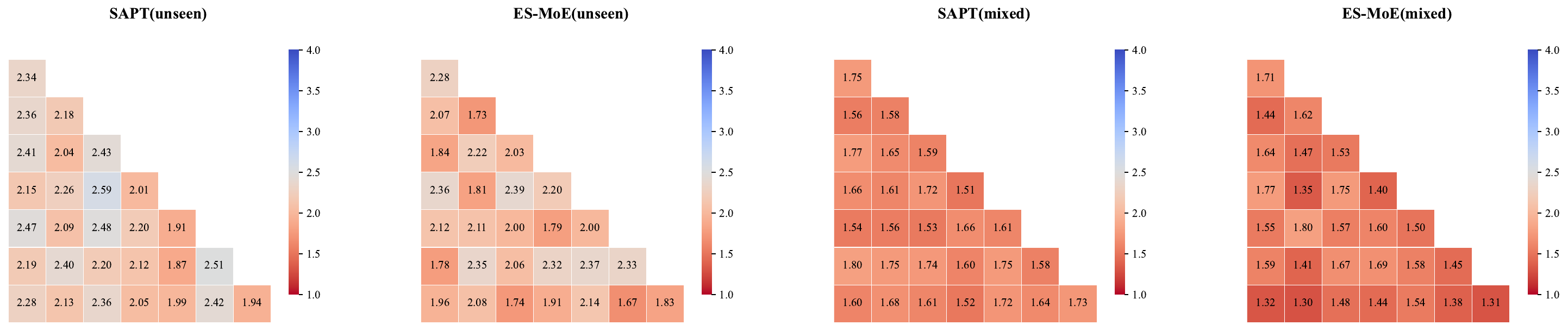}
    \caption{Detailed AF results of four approaches on L$^2$-EMG (Unseen), L$^2$-EMG (Mixed), where bluer represents severe forgetting of generation ability across different motion scenarios, indicating low performance; redder represents strong retention of generation ability across different motion scenarios, indicating high performance.}
    \label{fig:vis}
\end{figure*}

\begin{figure*}
    \centering
    \includegraphics[width=\linewidth]{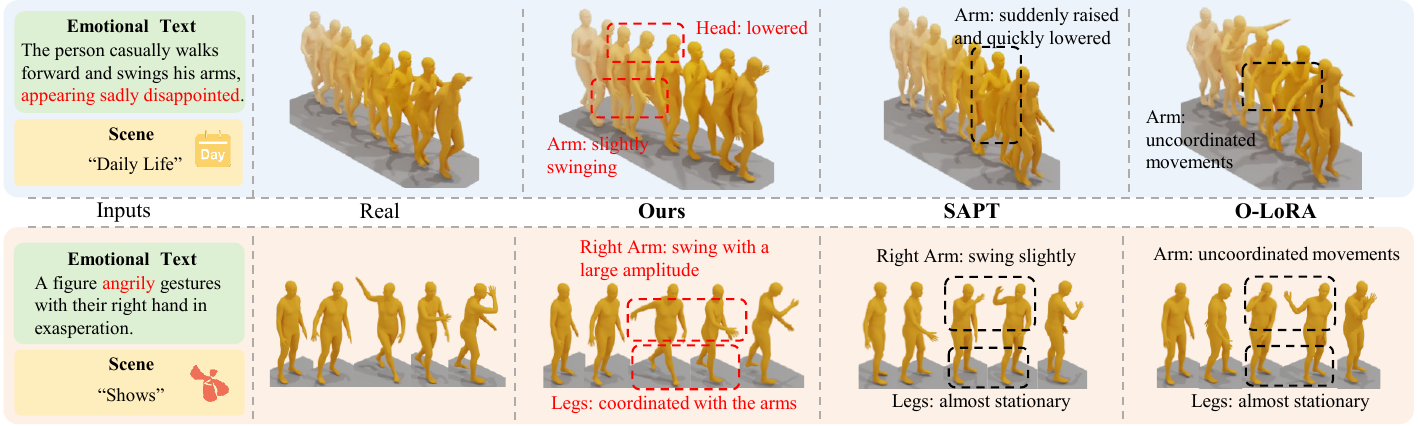}
    \caption{Visualization of motions generated by ES-MoE and other baselines. Red boxes indicate more precise emotional expressions achieved by ES-MoE, whereas black boxes indicate the limitations observed in motions generated by the baselines.}
    \label{fig:qualititave comparision}
\end{figure*}

\section{Results and Discussion}

\textbf{Main Experimental Results.} Table \ref{tab:result} presents a comparative analysis of different approaches on the L$^2$-EMG dataset. From this table, we can see that: \textbf{1)} \textbf{Performance on Unseen L$^2$-EMG Dataset.} On the Unseen L$^2$-EMG Dataset, our ES-MoE approach outperforms other baselines on almost all metrics. For instance, compared to the best-performing baseline, SAPT, ES-MoE achieves better results on the AF, AR, and AD metrics, suggesting that it generates more natural, coherent, and text-aligned motions in diverse scenarios. On the AMM metric, ES-MoE achieves comparable results, indicating that the motions generated by ES-MoE are diverse and rich. Moreover, ES-MoE also achieves better performance in the AWF and FR metrics, further confirming its effectiveness in decoupling emotional information from motions and its flexibility in adapting to entirely new motion scenarios.
\textbf{2)} \textbf{Performance on Mixed L$^2$-EMG Dataset.} On the Mixed L$^2$-EMG Dataset which better aligns with real-world applications, our ES-MoE approach also surpasses other baselines on most metrics. This indicates that ES-MoE is not only effective at decoupling emotional information and generating realistic and natural emotional motions but also adapts well to the real-world scenario.

\textbf{Effectiveness Study for Emotion Decoupling.} To validate ES-MoE’s effectiveness in addressing the emotion decoupling challenge, we conduct ablation studies (Table \ref{tab:result}). Specifically, \textbf{w/o CGED} and \textbf{w/o $\mathcal{L}_{emo}$} denote settings where the Causal-Guided Emotion Decoupling Block (CGED) and the $\mathcal{L}_{emo}$ do not work, respectively. From this table, we can see that: 1) \textbf{w/o CGED} performs worse on both unseen and mixed datasets, especially in AWF score. This justifies the effectiveness of the CGED block in decoupling emotional commonality and enabling the generation of more emotionally expressive motions. 2) \textbf{w/o $\mathcal{L}_{emo}$} also shows a performance drop in AWF and AR on both datasets. This further demonstrates the effectiveness of the CGED block in addressing the challenge of emotional decoupling.

\textbf{Effectiveness Study for Scenario Adapting.} To validate the capability of ES-MoE in addressing the scenario adaptation challenge, we conduct related experiments, as shown in Table 2. Specifically, \textbf{w/o SAMoE} refers to the setting where the Scenario-Adaptive Mixture of Experts(SAMoE) does not work. From this table, we can see that: \textbf{w/o SAMoE} shows a significant performance drop on both types of datasets compared to ES-MoE, with all metrics substantially decreased, especially AF and AR. This demonstrates that scenario-adapted experts can efficiently adapt to newly introduced motion scenarios, justifying the effectiveness of SAMoE in addressing the scenario adapting challenge.

\textbf{Forgetting Analysis across Different Scenarios.} We visualize the FID performance of SAPT (best baseline) and ES-MoE on Unseen and Mixed L$^2$-EMG datasets in Figure \ref{fig:vis}. Each matrix entry (i, j) denotes the FID score on the j-th scenario after training on the i-th scenario. Results show: 1) Final-scenario tests show no clear performance drop—sometimes improvement—indicating minimal forgetting in cross-scenario lifelong learning. 2) SAPT and ES-MoE achieve effective knowledge transfer, with ES-MoE consistently yielding lower FID, reflecting more accurate emotion decoupling and stronger scenario adaptability.

\textbf{Qualitative Analysis via Visualization.} Figure \ref{fig:qualititave comparision} shows visual comparisons of motions generated by ES-MoE and other methods. We observe: 1) All methods capture basic patterns such as walking and arm swinging, but O-LoRA often produces incoherent or uncoordinated motions—for example, in the first case, it swings both arms while walking, causing unnatural movement. 2) SAPT shows limited emotional understanding. In the second case, although it generates a coherent waving motion, the amplitude is too small to convey the intended angry emotion. In contrast, ES-MoE produces coordinated motions that accurately express the target emotion and match the scenario.

\section{Conclusion}
In this paper, we propose a new and challenging \textbf{L}LM-Centric \textbf{L}ifelong \textbf{E}mpathic \textbf{M}otion \textbf{G}eneration (L$^2$-EMG) task aimed at enhancing the lifelong learning ability of existing motion generation models in unseen scenarios. To address the L$^2$-EMG task, we propose an \textbf{E}motion-Transferable and \textbf{S}cenario-Adapted \textbf{M}ixture \textbf{o}f \textbf{E}xperts (ES-MoE) approach. The ES-MoE method consists of a causal-guided emotion decoupling block and a scenario-adapted expert constructing block, designed to tackle the challenges of the sustainable transfer of common emotional representations and the non-forgetting of scenario-specific motion characteristics, respectively. To comprehensively evaluate the ES-MoE approach, we construct multiple L$^2$-EMG datasets. Experimental results on the L$^2$-EMG dataset demonstrate the superior performance of ES-MoE compared to several state-of-the-art baselines. In our future work, we would like to transfer our ES-MoE approach to other tasks across diverse scenarios, e.g., human-scenario interaction motion generation \cite{interact,4d}, where scenario continual adaptation remains a key challenge in this task. Additionally, we also would like to use emotional motions as conditional guidance to assist humanoid robot control \cite{humanrobot}, in order to empower these humanoid robots with not only intelligence but also empathetic ability.

\section{Acknowledgments}

This work was supported by two NSFC grants, i.e., No. 62576234, No.62376178 and sponsored by CIPS-LMG Huawei Innovation Fund. This work was also supported by Collaborative Innovation Center of Novel Software Technology and Industrialization, and a Project Funded by the Priority Academic Program Development of Jiangsu Higher Education Institutions (PAPD).

\bibliography{Formatting-Instructions-LaTeX-2026}

@inproceedings{DBLP:conf/mm/GuoZWZSDG020,
  author       = {Chuan Guo and
                  Xinxin Zuo and
                  Sen Wang and
                  Shihao Zou and
                  Qingyao Sun and
                  Annan Deng and
                  Minglun Gong and
                  Li Cheng},
  title        = {Action2Motion: Conditioned Generation of 3D Human Motions},
  booktitle    = {Proceedings of {MM} 2020},
  year         = {2020}
}

@article{DBLP:journals/pami/Liu0JJLL023,
  author       = {Zhenguang Liu and
                  Shuang Wu and
                  Shuyuan Jin and
                  Shouling Ji and
                  Qi Liu and
                  Shijian Lu and
                  Li Cheng},
  title        = {Investigating Pose Representations and Motion Contexts Modeling for
                  3D Motion Prediction},
  journal      = {{IEEE} Trans. Pattern Anal. Mach. Intell.},
  volume       = {45},
  number       = {1},
  pages        = {681--697},
  year         = {2023}
}

@inproceedings{DBLP:conf/iccv/MaoLSL19,
  author       = {Wei Mao and
                  Miaomiao Liu and
                  Mathieu Salzmann and
                  Hongdong Li},
  title        = {Learning Trajectory Dependencies for Human Motion Prediction},
  booktitle    = {Proceedings of {ICCV} 2019},
  year         = {2019}
}

@inproceedings{DBLP:conf/cvpr/GuoZZ0JL022,
  author       = {Chuan Guo and
                  Shihao Zou and
                  Xinxin Zuo and
                  Sen Wang and
                  Wei Ji and
                  Xingyu Li and
                  Li Cheng},
  title        = {Generating Diverse and Natural 3D Human Motions from Text},
  booktitle    = {Proceedings of {CVPR} 2022},
  year         = {2022}
}

@inproceedings{DBLP:conf/eccv/TevetGHBC22,
  author       = {Guy Tevet and
                  Brian Gordon and
                  Amir Hertz and
                  Amit H. Bermano and
                  Daniel Cohen{-}Or},
  title        = {MotionCLIP: Exposing Human Motion Generation to {CLIP} Space},
  booktitle    = {Proceedings of {ECCV} 2022},
  year         = {2022}
}

@article{DBLP:journals/pami/ZhangCPHGYL24,
  author       = {Mingyuan Zhang and
                  Zhongang Cai and
                  Liang Pan and
                  Fangzhou Hong and
                  Xinying Guo and
                  Lei Yang and
                  Ziwei Liu},
  title        = {MotionDiffuse: Text-Driven Human Motion Generation With Diffusion
                  Model},
  journal      = {{IEEE} Trans. Pattern Anal. Mach. Intell.},
  volume       = {46},
  number       = {6},
  pages        = {4115--4128},
  year         = {2024}
}

@inproceedings{DBLP:conf/iclr/TevetRGSCB23,
  author       = {Guy Tevet and
                  Sigal Raab and
                  Brian Gordon and
                  Yonatan Shafir and
                  Daniel Cohen{-}Or and
                  Amit Haim Bermano},
  title        = {Human Motion Diffusion Model},
  booktitle    = {Proceedings of {ICLR} 2023},
  year         = {2023}
}

@inproceedings{DBLP:conf/iccv/ZhangGPCHLYL23,
  author       = {Mingyuan Zhang and
                  Xinying Guo and
                  Liang Pan and
                  Zhongang Cai and
                  Fangzhou Hong and
                  Huirong Li and
                  Lei Yang and
                  Ziwei Liu},
  title        = {ReMoDiffuse: Retrieval-Augmented Motion Diffusion Model},
  booktitle    = {Proceedings of{ICCV} 2023},
  year         = {2023}
}

@article{abs-2301-06052,
  author       = {Jianrong Zhang and
                  Yangsong Zhang and
                  Xiaodong Cun and
                  Shaoli Huang and
                  Yong Zhang and
                  Hongwei Zhao and
                  Hongtao Lu and
                  Xi Shen},
  title        = {{T2M-GPT:} Generating Human Motion from Textual Descriptions with
                  Discrete Representations},
  journal      = {CoRR},
  volume       = {abs/2301.06052},
  year         = {2023}
}

@article{DBLP:journals/corr/abs-2312-00063,
  author       = {Chuan Guo and
                  Yuxuan Mu and
                  Muhammad Gohar Javed and
                  Sen Wang and
                  Li Cheng},
  title        = {MoMask: Generative Masked Modeling of 3D Human Motions},
  journal      = {CoRR},
  volume       = {abs/2312.00063},
  year         = {2023}
}

@inproceedings{DBLP:conf/aaai/ZhangHLTLC00YO24,
  author       = {Yaqi Zhang and
                  Di Huang and
                  Bin Liu and
                  Shixiang Tang and
                  Yan Lu and
                  Lu Chen and
                  Lei Bai and
                  Qi Chu and
                  Nenghai Yu and
                  Wanli Ouyang},
  title        = {MotionGPT: Finetuned LLMs Are General-Purpose Motion Generators},
  booktitle    = {Proceedings of {AAAI} 2024},
  year         = {2024}
}

@inproceedings{DBLP:conf/cvpr/ChenJLHFCY23,
  author       = {Xin Chen and
                  Biao Jiang and
                  Wen Liu and
                  Zilong Huang and
                  Bin Fu and
                  Tao Chen and
                  Gang Yu},
  title        = {Executing your Commands via Motion Diffusion in Latent Space},
  booktitle    = {Proceedings of {CVPR} 2023},
  year         = {2023}
}

@inproceedings{DBLP:conf/nips/OordVK17,
  author       = {Oord},
  title        = {Neural Discrete Representation Learning},
  booktitle    = {Proceedings of {NeurIPS} 2017},
  year         = {2017}
}

@inproceedings{DBLP:conf/iclr/HuSWALWWC22,
  author       = {Edward J. Hu and
                  Yelong Shen and
                  Phillip Wallis and
                  Zeyuan Allen{-}Zhu and
                  Yuanzhi Li and
                  Shean Wang and
                  Lu Wang and
                  Weizhu Chen},
  title        = {LoRA: Low-Rank Adaptation of Large Language Models},
  booktitle    = {Proceedings of {ICLR} 2022},
  year         = {2022}
}

@inproceedings{DBLP:conf/iclr/ZengLDWL0YXZXTM23,
  author       = {Aohan Zeng and
                  Xiao Liu and
                  Zhengxiao Du and
                  Zihan Wang and
                  Hanyu Lai and
                  Ming Ding and
                  Zhuoyi Yang and
                  Yifan Xu and
                  Wendi Zheng and
                  Xiao Xia and
                  Weng Lam Tam and
                  Zixuan Ma and
                  Yufei Xue and
                  Jidong Zhai and
                  Wenguang Chen and
                  Zhiyuan Liu and
                  Peng Zhang and
                  Yuxiao Dong and
                  Jie Tang},
  title        = {{GLM-130B:} An Open Bilingual Pre-trained Model},
  booktitle    = {Proceedings of {ICLR} 2023},
  year         = {2023}
}

@inproceedings{DBLP:conf/nips/LinZLCZWZ23,
  author       = {Jing Lin and
                  Ailing Zeng and
                  Shunlin Lu and
                  Yuanhao Cai and
                  Ruimao Zhang and
                  Haoqian Wang and
                  Lei Zhang},
  title        = {Motion-X: {A} Large-scale 3D Expressive Whole-body Human Motion Dataset},
  booktitle    = {Proceedings of {NeurIPS} 2023},
  year         = {2023}
}

@inproceedings{DBLP:conf/iccv/GhoshCOTS21,
  author       = {Anindita Ghosh and
                  Noshaba Cheema and
                  Cennet Oguz and
                  Christian Theobalt and
                  Philipp Slusallek},
  title        = {Synthesis of Compositional Animations from Textual Descriptions},
  booktitle    = {Proceedings of {ICCV} 2021},
  year         = {2021}
}

@article{DBLP:journals/neco/JacobsJNH91,
  author       = {Robert A. Jacobs and
                  Michael I. Jordan and
                  Steven J. Nowlan and
                  Geoffrey E. Hinton},
  title        = {Adaptive Mixtures of Local Experts},
  journal      = {Neural Comput.},
  volume       = {3},
  number       = {1},
  pages        = {79--87},
  year         = {1991}
}

@inproceedings{DBLP:conf/cvpr/TsengCL23,
  author       = {Jonathan Tseng and dsaad and sdsadad and sadadassd and dasda},
  title        = {{EDGE:} Editable Dance Generation From Music},
  booktitle    = {Proceedings of {CVPR} 2023},
  year         = {2023}
}

@inproceedings{DBLP:conf/aaai/LiZZS22,
  author       = {Buyu Li and
                  Yongchi Zhao and
                  Zhelun Shi and
                  Lu Sheng},
  title        = {DanceFormer: Music Conditioned 3D Dance Generation with Parametric
                  Motion Transformer},
  booktitle    = {Proceedings of {AAAI} 2022},
  year         = {2022}
}

@inproceedings{JiangCLYYC23,
  author       = {Biao Jiang and
                  Xin Chen and
                  Wen Liu and
                  Jingyi Yu and
                  Gang Yu and
                  Tao Chen},
  title        = {MotionGPT: Human Motion as a Foreign Language},
  booktitle    = {Proceedings of {NeurIPS} 2023},
  year         = {2023}
}

@article{yang2024unimumo,
  title={UniMuMo: Unified Text, Music and Motion Generation},
  author={Yang, Han and Su, Kun and Zhang, Yutong and Chen, Jiaben and Qian, Kaizhi and Liu, Gaowen and Gan, Chuang},
  journal={arXiv preprint arXiv:2410.04534},
  year={2024}
}

@inproceedings{DBLP:conf/aaai/IseleC18,
  author       = {David Isele and
                  Akansel Cosgun},
  title        = {Selective Experience Replay for Lifelong Learning},
  booktitle    = {Proceedings of {AAAI} 2018},
  year         = {2018},
}

@article{DBLP:journals/corr/KirkpatrickPRVD16,
  author       = {James Kirkpatrick and
                  Razvan Pascanu and
                  Neil C. Rabinowitz and
                  Joel Veness and
                  Guillaume Desjardins and
                  Andrei A. Rusu and
                  Kieran Milan and
                  John Quan and
                  Tiago Ramalho and
                  Agnieszka Grabska{-}Barwinska and
                  Demis Hassabis and
                  Claudia Clopath and
                  Dharshan Kumaran and
                  Raia Hadsell},
  title        = {Overcoming catastrophic forgetting in neural networks},
  journal      = {CoRR},
  volume       = {abs/1612.00796},
  year         = {2016},
}

@article{posegpt,
  author       = {Yao Feng and
                  Jing Lin and
                  Sai Kumar Dwivedi and
                  Yu Sun and
                  Priyanka Patel and
                  Michael J. Black},
  title        = {PoseGPT: Chatting about 3D Human Pose},
  journal      = {CoRR},
  volume       = {abs/2311.18836},
  year         = {2023},
}

@inproceedings{tyumm2024,
  author       = {Tan Yu and
                  Jingjing Wang and
                  Jiawen Wang and
                  Jiamin Luo and
                  Guodong Zhou},
  title        = {Towards Emotion-enriched Text-to-Motion Generation via LLM-guided
                  Limb-level Emotion Manipulating},
  booktitle    = {Proceedings of {ACM} {MM} 2024},
  year         = {2024},
}

@article{lifeifei,
  author       = {Changan Chen and
                  Juze Zhang and
                  Shrinidhi K. Lakshmikanth and
                  Yusu Fang and
                  Ruizhi Shao and
                  Gordon Wetzstein and
                  Li Fei{-}Fei and
                  Ehsan Adeli},
  title        = {The Language of Motion: Unifying Verbal and Non-verbal Language of
                  3D Human Motion},
  journal      = {CoRR},
  volume       = {abs/2412.10523},
  year         = {2024},
}

@inproceedings{DBLP:conf/iclr/PengT0YJ24,
  author       = {Bohao Peng and
                  Zhuotao Tian and
                  Shu Liu and
                  Ming{-}Chang Yang and
                  Jiaya Jia},
  title        = {Scalable Language Model with Generalized Continual Learning},
  booktitle    = {Proceedings of {ICLR} 2024},
  year         = {2024},
}

@inproceedings{DBLP:journals/corr/SaxeMG13,
  author       = {Andrew M. Saxe and sddfs and sadsdsd and sdfsfs and dsdfsfd},
  title        = {Exact solutions to the nonlinear dynamics of learning in deep linear
                  neural networks},
  booktitle    = {Proceedings of {ICLR} 2014},
  year         = {2014},
}

@inproceedings{sapt,
  author       = {Weixiang Zhao and
                  Shilong Wang and
                  Yulin Hu and
                  Yanyan Zhao and
                  Bing Qin and
                  Xuanyu Zhang and
                  Qing Yang and
                  Dongliang Xu and
                  Wanxiang Che},
  title        = {{SAPT:} {A} Shared Attention Framework for Parameter-Efficient Continual
                  Learning of Large Language Models},
  booktitle    = {Proceedings of {ACL} 2024},
  year         = {2024},
}

@inproceedings{DBLP:conf/nips/Lopez-PazR17,
  author       = {David Lopez{-}Paz and
                  Marc'Aurelio Ranzato},
  title        = {Gradient Episodic Memory for Continual Learning},
  booktitle    = {Proceedings of {NeurIPS} 2017},
  year         = {2017},
}

@article{mola,
  author       = {Chongyang Gao and
                  Kezhen Chen and
                  Jinmeng Rao and
                  Baochen Sun and
                  Ruibo Liu and
                  Daiyi Peng and
                  Yawen Zhang and
                  Xiaoyuan Guo and
                  Jie Yang and
                  V. S. Subrahmanian},
  title        = {Higher Layers Need More LoRA Experts},
  journal      = {CoRR},
  volume       = {abs/2402.08562},
  year         = {2024},
}

@inproceedings{o-lora,
  author       = {Xiao Wang and
                  Tianze Chen and
                  Qiming Ge and
                  Han Xia and
                  Rong Bao and
                  Rui Zheng and
                  Qi Zhang and
                  Tao Gui and
                  Xuanjing Huang},
  title        = {Orthogonal Subspace Learning for Language Model Continual Learning},
  booktitle    = {Proceedings of {EMNLP} 2023},
  year         = {2023},
}

@inproceedings{epi,
  author       = {Zhicheng Wang and
                  Yufang Liu and
                  Tao Ji and
                  Xiaoling Wang and
                  Yuanbin Wu and
                  Congcong Jiang and
                  Ye Chao and
                  Zhencong Han and
                  Ling Wang and
                  Xu Shao and
                  Wenqiu Zeng},
  title        = {Rehearsal-free Continual Language Learning via Efficient Parameter
                  Isolation},
  booktitle    = {Proceedings of {ACL} 2023},
  year         = {2023},
}

@inproceedings{pp,
  author       = {Anastasia Razdaibiedina and
                  Yuning Mao and
                  Rui Hou and
                  Madian Khabsa and
                  Mike Lewis and
                  Amjad Almahairi},
  title        = {Progressive Prompts: Continual Learning for Language Models},
  booktitle    = {Proceedings of {ICLR} 2023},
  year         = {2023},
}

@article{lwf,
  author       = {Zhizhong Li 
                  and sdaads and sdasdadas and sdadadad and adsasdad},
  title        = {Learning without Forgetting},
  journal      = {{IEEE} Trans. Pattern Anal. Mach. Intell.},
  volume       = {40},
  number       = {12},
  pages        = {2935--2947},
  year         = {2018},
}

@article{humanrobot,
  author       = {Jiageng Mao and
                  Siheng Zhao and
                  Siqi Song and
                  Tianheng Shi and
                  Junjie Ye and
                  Mingtong Zhang and
                  Haoran Geng and
                  Jitendra Malik and
                  Vitor Guizilini and
                  Yue Wang},
  title        = {Learning from Massive Human Videos for Universal Humanoid Pose Control},
  journal      = {CoRR},
  volume       = {abs/2412.14172},
  year         = {2024},
}

@inproceedings{interact,
  author       = {Nan Jiang and
                  Zimo He and
                  Zi Wang and
                  Hongjie Li and
                  Yixin Chen and
                  Siyuan Huang and
                  Yixin Zhu},
  title        = {Autonomous Character-Scene Interaction Synthesis from Text Instruction},
  booktitle    = {Proceedings of {SIGGRAPH} 2024},
  year         = {2024},
}

@book{book,
  author = {Judea Pearl and sdajnjad and asdad and sadadadad and asdada}, 
  title = {The book of why: the new science of cause and effect},
  year = 2018,
}

@article{dropout,
  author       = {Nitish Srivastava and
                  Geoffrey E. Hinton and
                  Alex Krizhevsky and
                  Ilya Sutskever and
                  Ruslan Salakhutdinov},
  title        = {Dropout: a simple way to prevent neural networks from overfitting},
  journal      = {J. Mach. Learn. Res.},
  volume       = {15},
  year         = {2014},
}

@inproceedings{show,
  author       = {Kelvin Xu and
                  Jimmy Ba and
                  Ryan Kiros and
                  Kyunghyun Cho and
                  Aaron C. Courville and
                  Ruslan Salakhutdinov and
                  Richard S. Zemel and
                  Yoshua Bengio},
  title        = {Show, Attend and Tell: Neural Image Caption Generation with Visual
                  Attention},
  booktitle    = {Proceedings of {ICML} 2015},
  year         = {2015},
}

@article{kmeans,
  title={Algorithm AS 136: A K-Means Clustering Algorithm},
  author={ Wong, J. A. Hartiganm. A. },
  journal={Journal of the Royal Statistical Society},
  volume={28},
  year={1979},
}

@article{4d,
  author       = {Hongjie Li and
                  Hong{-}Xing Yu and
                  Jiaman Li and
                  Jiajun Wu},
  title        = {ZeroHSI: Zero-Shot 4D Human-Scene Interaction by Video Generation},
  journal      = {CoRR},
  volume       = {abs/2412.18600},
  year         = {2024},
}

@inproceedings{boost,
  author       = {Jiazuo Yu and
                  Yunzhi Zhuge and
                  Lu Zhang and
                  Ping Hu and
                  Dong Wang and
                  Huchuan Lu and
                  You He},
  title        = {Boosting Continual Learning of Vision-Language Models via Mixture-of-Experts
                  Adapters},
  booktitle    = {Proceedings of {CVPR} 2024},
  year         = {2024},
}

@article{casual1,
  author       = {Xiangmeng Wang and
                  Qian Li and
                  Dianer Yu and
                  Peng Cui and
                  Zhichao Wang and
                  Guandong Xu},
  title        = {Causal Disentanglement for Semantics-Aware Intent Learning in Recommendation},
  journal      = {CoRR},
  year         = {2022},
}

@inproceedings{casual2,
  author       = {Yang Liu and
                  Zhaoyang Xia and
                  Mengyang Zhao and
                  Donglai Wei and
                  Yuzheng Wang and
                  Siao Liu and
                  Bobo Ju and
                  Gaoyun Fang and
                  Jing Liu and
                  Liang Song},
  title        = {Learning Causality-inspired Representation Consistency for Video Anomaly
                  Detection},
  booktitle    = {Proceedings of the 31st {ACM} International Conference on Multimedia,
                  {MM} 2023, Ottawa, ON, Canada, 29 October 2023- 3 November 2023},
  year         = {2023},
}

@ARTICLE{zh,
  author={Zhang, Han and Wang, Jingjing and Luo, Jiamin and Zhang, Min and Zhou, Guodong},
  journal={IEEE Transactions on Audio, Speech and Language Processing}, 
  title={Boosting LLM’s Continual Sentiment Understanding for Low-Resource Languages}, 
  year={2025},
  volume={33},
 }

\end{document}